\documentclass[letterpaper, 10pt, conference]{ieeeconf}
\IEEEoverridecommandlockouts	

\usepackage{color,soul}
\usepackage{multirow}
\usepackage{lipsum}
\usepackage{amsmath}
\usepackage{amssymb}
\usepackage[ruled,vlined]{algorithm2e}
\usepackage{graphicx}
\usepackage{multirow}
\usepackage[normalem]{ulem}
\useunder{\uline}{\ul}{}
\graphicspath{{./figures/}}
\usepackage{siunitx}
\usepackage{tabularx, booktabs}
\usepackage[para,online,flushleft]{threeparttable}
\usepackage[caption=false,font=normalsize,labelfont=sf,textfont=sf]{subfig}
\title{\LARGE \bf BirdNet: a 3D Object Detection Framework \\ from LiDAR information}

\author{Jorge Beltr\'{a}n, Carlos Guindel, Francisco Miguel Moreno, \\ Daniel Cruzado, Fernando Garc\'{i}a, \IEEEmembership{Member, IEEE} and Arturo de la Escalera  \\ Intelligent Systems Laboratory (LSI) Research Group\\
Universidad Carlos III de Madrid, Legan\'{e}s, Madrid, Spain\\
\{jbeltran, cguindel, franmore, dcruzado, fegarcia, escalera\}@ing.uc3m.es
}
\begin{document} 

\maketitle
\thispagestyle{empty}
\pagestyle{empty}

\begin{abstract}
Understanding driving situations regardless the conditions of the traffic scene is a cornerstone on the path towards autonomous vehicles; however, despite common sensor setups already include complementary devices such as LiDAR or radar, most of the research on perception systems has traditionally focused on computer vision. We present a LiDAR-based 3D object detection pipeline entailing three stages. First, laser information is projected into a novel cell encoding for bird's eye view projection. Later, both object location on the plane and its heading are estimated through a convolutional neural network originally designed for image processing. Finally, 3D oriented detections are computed in a post-processing phase. Experiments on KITTI dataset show that the proposed framework achieves state-of-the-art results among comparable methods. Further tests with different LiDAR sensors in real scenarios assess the multi-device capabilities of the approach.
\end{abstract}


%
\IEEEpeerreviewmaketitle

\section{Introduction}
In recent years, the rapid evolution of technology has significantly boosted the level of automation in road vehicles. In a brief period of time, cars manufacturers have gone from advertising Advanced Driver Assistance Systems (ADAS) or automatic parking maneuvers to commercial vehicles offering a high degree of automation in roads. This progress would not have been possible without the growth of artificial intelligence algorithms, which permit extracting knowledge from large amounts of real driving data and, thus, developing more robust control and perception systems.

However, to switch from the current automation level to actual self-driving cars, further developments in perception systems are still required in order to enable full scene understanding so that the best driving decisions can be made.

In this regard, object detection algorithms will play a major role, as they constitute a key link in the task of identification and prediction of potential hazards in the roads.
 
Lately, research on object detection for autonomous driving has mainly focused on image understanding tasks, probably motivated by the great progress experienced by machine learning and, especially, deep learning techniques applied to computer vision, where they have shown their ability to handle the scene variations commonly found in driving scenarios, such as different object orientations, changes in illumination and occlusions, among others.

Nonetheless, automated vehicles usually mount other kinds of sensors capable of perceiving spatial data, thus increasing the robustness of the system. LiDAR sensors complement cameras information and work under conditions where computer vision often finds difficulties, such as darkness, fog or rain.

\begin{figure*}[htb]
\label{fig:overview}
\centering
\includegraphics[width=\linewidth]{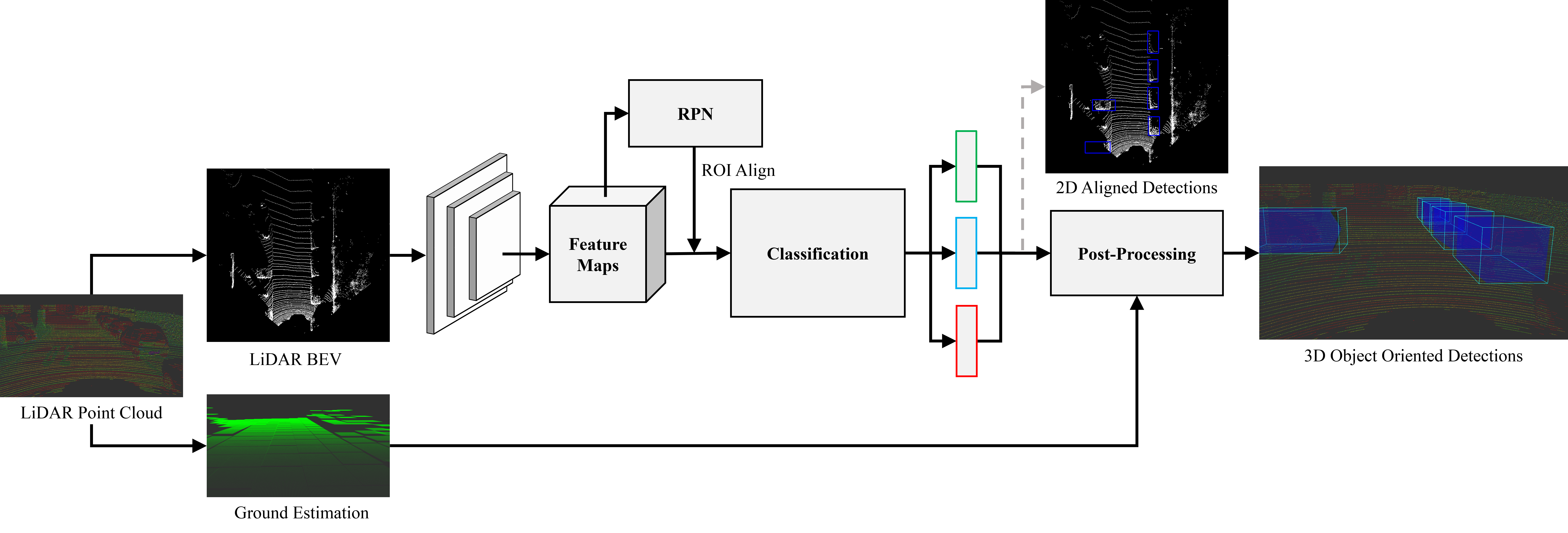}
\caption{BirdNet 3D object detection framework. The three outputs of the network are: class (green), 2d bounding box (blue) and yaw angle (red).}
\end{figure*}

In order to fill this gap and provide a functional and robust perception system even when camera data is not reliable due to scene conditions, we present BirdNet, a 3D detection and classification method based on LiDAR information. To comply with real-time requirements, the proposed approach is based on a state-of-the-art detector \cite{Ren2015}. To be fed into the network, the LiDAR point cloud is encoded as a bird's eye view (BEV) image as explained in Sec. \ref{sec:bev}, minimizing the information loss produced by the projection. In the last stage, the 2D detections coming from the network are then processed in conjunction with the BEV to obtain the final 3D bounding boxes of the obstacles. A general overview of the framework is shown in Fig. \ref{fig:overview}. 





In order to assess the performance of the proposed approach, we evaluate our results on the tasks of 2D detection, bird's eye view (BEV) and 3D detection on the challenging KITTI Object Detection Benchmark \cite{Geiger2012KITTI}.

The main contributions of the paper are:
\begin{itemize}
\item A novel cell encoding proposal for BEV, invariant to distance and differences on LiDAR devices resolution.
\item A 3D detection framework capable of identifying cars, cyclists, and pedestrians taking a BEV image as input.
\end{itemize}
The rest of this paper is organized as follows. In Section II, a brief review of related work is provided. In Section III, a detailed description of the proposed approach is presented. Experimental results are discussed in Section IV. Finally, the conclusions of the paper are drawn in Section V.

\section{Related Work}
\label{sec:related}
Research on autonomous perception has commonly focused on modeling advanced hand-crafted features, such as HoG \cite{Dalal2005} or Haar \cite{Viola2001}. 

Nonetheless, since the emergence of modern Convolutional Neural Networks (CNNs) and large-scale image datasets such as ImageNet \cite{Russakovsky2015ImageNet}, object detection studies have progressively moved towards feature learning approaches, which produce more robust representations of objects, increasing the performance on classification tasks. 

Although CNNs were originally applied to image classification, approaches aiming to integrate them into a full object detection pipeline shortly became popular. Firstly, by using external segmented objects \cite{Girshick2015a}, and later by developing the Region Proposals Networks (RPNs) \cite{Ren2015}.

In the latest years, on-board 3D object detection task has become a key component in the development of autonomous vehicles. Given the most common vehicle sensor setups, works have been focused on three main research lines.

\textit{Monocular 3D.}
Besides 2D detections, some approaches have been made attempting to provide spatial location of objects based on visual information. In 
\cite{Chen16}, 3D object candidates are placed over a ground plane prior and classified in image space. Similarly, 3D voxel patterns have been used to estimate position and orientation of objects in the scene \cite{Xiang15}.

\textit{Point cloud.} 
Some other works have used point cloud data to compute object detections in 3D, either using information from stereo cameras or laser sensors. Although there are some methods which use hand-crafted features \cite{3DOPChen2015} \cite{Babahajiani2015}, latest approaches take advantage of feature learning capabilities. 

Among these latter group, two different strategies are being explored. On the one hand, some approaches work with spatial information by turning the 3D space into a voxel grid and applying 3D convolutions \cite{Engelcke2016} \cite{3DFCNli2017} \cite{Zhou2017}. On the other hand, 2D CNNs are used by projecting LiDAR point cloud into a front view \cite{Li16VeloFCN} or a bird's eye view (BEV) \cite{Yu2017}.

\textit{Data fusion.}
Since camera and LiDAR are complementary data sources, many works have tried to build robust object detection frameworks by fusing their information. Traditional methods used to obtain 3D ROIs from the point cloud and perform classification on their image projection \cite{Nando2017}. However, when RPNs step in, the candidate proposal stage was outperformed both in ROIs quality and execution time. 
In this direction, a novel approach has been presented in \cite{Qi2017} where regions of interest and classification are computed on the image space, and final location is performed over the LiDAR data.


Our work falls into the category of point cloud-based 3D object detection. Nonetheless, in order to meet the time requirements of autonomous perception systems, a 2D image detector is used. For that purpose, LiDAR information is projected into their BEV representation.

\section{Proposed Approach}
\label{sec:proposed}

\subsection{Bird eye's view generation}
\label{sec:bev}
\subsubsection{3D point cloud representation} 
\hfill\\
The data acquired from the LiDAR sensor is provided as a point cloud, including the coordinates in meters and the intensity value from the reflected laser beam. This information is converted to a BEV image, with total size $N \times N$ meters and cell size $\delta$, which is a 3-channel image encoding height, intensity, and density information. Firstly, the height channel represents the maximum height given by a point at each cell, and it is limited to a maximum $H_\text{top}$ of 3 meters above the ground. Secondly, the intensity channel encodes the mean intensity of all points that lay in the cell. Finally, the last channel represents the density of points in each cell, which is computed as the number of points in that cell, divided by the maximum possible number of points. This normalization process is performed in order to obtain homogeneous and meaningful values among all cells, as it is described below.

\subsubsection{Density normalization}
\hfill\\
The principal difference between LiDAR sensors lies in the number of points they collect from the environment, which is mainly due to the number of vertical planes they have, $N_p$, and their horizontal resolution, $\Delta\theta$. This difference between sensors is reflected in the bird eye's view as a big variation in the density channel. As a result, DNNs trained with data coming from a certain sensor cannot be used with different LiDAR devices.


In order to deal with this problem, a normalization map is proposed to take into account the differences produced by the amount of lasers beams projected by the LiDAR. This map consists of a 2D grid with the same size and resolution as the bird eye's view image, and it represents the maximum number of points each cell could contain in the best-case scenario. For this, the whole cell is considered a solid object with size $\delta \times \delta$ and height $H_\text{top}$, and the intersection between that object and the laser beams is analyzed.


The intersection analysis between each cell and the laser beams is performed plane by plane, where each LiDAR plane is actually a conical curve, due to the revolution of the sensor beam. Given this, if we consider the top view of the intersection, this one is simplified to the cut between a square and a circle, where there are three possible outcomes:

\begin{enumerate}
\item All square's vertexes are outside the circle.

\item All square's vertexes are inside the circle.

\item The circle intersects the square in two points $\lbrace P_1, P_2 \rbrace$.
\end{enumerate}

In the first case, the cell is not intersected by the LiDAR plane. In the second case, all points in that plane from angle $\theta_0$ to $\theta_n$ lay inside that cell, where $\theta_0$ is the horizontal angle of the first point in the cell that the sweep encounters, and $\theta_n$ is the last one. Finally, the third case will manifest when the plane passes through the top and bottom covers of the object, with height $H_{top}$ and $0$, respectively. In this case, the angles $\theta_0$ and $\theta_n$ are calculated as follows: First, only two segments of the square, $\lbrace CD, CB \rbrace$, are taken into account, where $C$ is the closest vertex to the sensor and $\lbrace B, D \rbrace$ are the vertexes connected to $C$. Then, the intersection points between the circle and those segments are calculated as follows, where $d$ is the distance from the sensor to the top (bottom) cover (the radius of the circle).


\begin{equation}\label{eq:cell_intersection_square_equation}
  \begin{cases}
    P_x^2 + P_y^2 = d^2 \\
    P_y - C_y = \frac{V_y - C_y}{V_x - C_x}(P_x - C_x), V \in \lbrace B,D \rbrace  \\
  \end{cases}
\end{equation}

The equation system for the circle and segments is shown in (\ref{eq:cell_intersection_square_equation}), from which the intersection point coordinates, $\lbrace P_x, P_y \rbrace$, are obtained. Here, $P$ is a generalization of the intersection points, $P_1$ and $P_2$, which are obtained from segments $CD$ and $CB$, respectively. This equation system will raise two possible solutions for each intersection point, from which only the point inside the segment $CV$ is valid. Finally, the input and output angles, $\theta_0$ and $\theta_n$, are calculated as $\theta = \arctan{({P_x}/{P_y})}$, where each intersection point will give one of the angles.



Finally, once the angles $\theta_0$ and $\theta_n$ are known, the number of points from plane $p$ that enter cell $\lbrace i,j \rbrace$ is computed using (\ref{eq:horizontal_points}) and the maximum number of points for that cell, $M_{i,j_{max}}$, is computed in (\ref{eq:max_points_cell}).

\begin{equation}\label{eq:horizontal_points}
  M_{p_{i,j}} = 
    \begin{cases}
      \lceil\frac{\left| \theta_n - \theta_0 \right|}{\Delta\theta}\rceil & \text{if plane intersects cell,} \\
      \multicolumn{1}{@{}c@{\quad}}{0} & \text{otherwise.}
    \end{cases}
\end{equation}
\begin{equation}\label{eq:max_points_cell}
  M_{i,j_{max}} = \sum_{p=0}^{N_p}{M_{p_{i,j}}}
\end{equation}

\subsection{Inference Framework} 
\subsubsection{Object Detection}
We adopt the Faster R-CNN meta-architecture \cite{Ren2015} to perform object detection on the previously generated multi-channel bird eye's view images. Faster R-CNN involves a two-stage process where feature maps generated by a feature-extractor CNN are used to generate proposals within the image, on the one hand, and to classify those proposals into different categories, on the other hand. Even though Faster R-CNN was designed to handle RGB images as an input, it can be viewed as a general framework that enables detection over arbitrary 2D structures such as the BEV images.

In this paper, the VGG-16 architecture \cite{Simonyan2015} is used as feature extractor. As usual, we sample features at the last convolutional layer of the \textit{backbone}; that is, \textit{conv5}. However, we noticed that the resolution of the resulting feature map, 16 times smaller than the input image, was not suitable for detection of instances whose representation in the BEV is limited to a few pixels, such as pedestrians and cyclists. Following \cite{Chen17}, we experiment removing the fourth max pooling layer, so that the downsampling performed throughout the feature extraction is reduced to 8. Besides, we adopt the ROIAlign method introduced in \cite{He2017} for feature pooling to increase the location accuracy.

We also use an appropriate set of RPN anchors, selected through a statistical analysis of the geometry of the projection of road users on the BEV. For efficiency reasons, only three scales, with box areas of $16^2$, $48^2$, and $80^3$ pixels, and three ratios, of $1:1$, $1:2$, and $2:1$, are employed. 


\subsubsection{Orientation Estimation}
In addition to the object detection task, we endow Faster R-CNN with the ability to detect the orientation of the objects. We adopt the approach described in \cite{Guindel2017ICVES} to that end. Objects' yaw angle is discretized, and a new sibling branch is added on top of the feature extractor and the common set of fully connected layers to perform class-aware multinomial classification into $N_b$ discrete angle bins. The estimated yaw angle is then obtained as the weighted average of the center of the predicted angle bin and its most probable neighbor; the probabilities of the respective bins, available through a softmax normalization, are used as weights. Although we let all bins be equal in size, we choose the origin so that the most general orientations (i.e., forward/rear, and left/right) are represented without error.

\begin{figure}[t]
\centering
\subfloat[]{\includegraphics[height=1.2in]{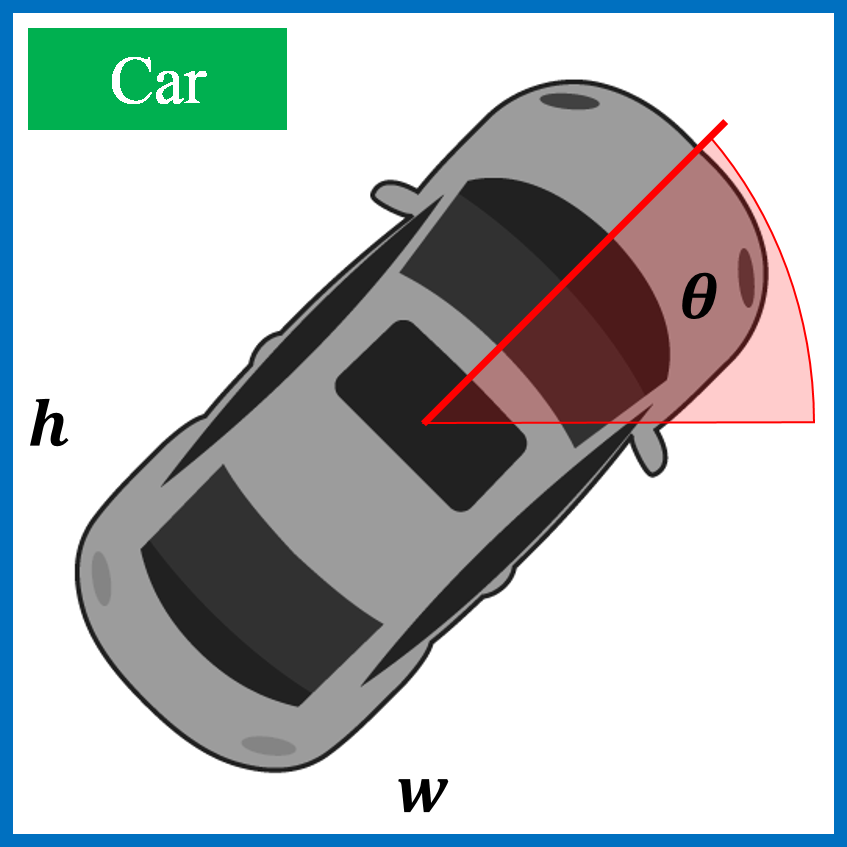}%
\label{fig:2ddetection}}
\hfil
\subfloat[]{\includegraphics[height=1.2in]{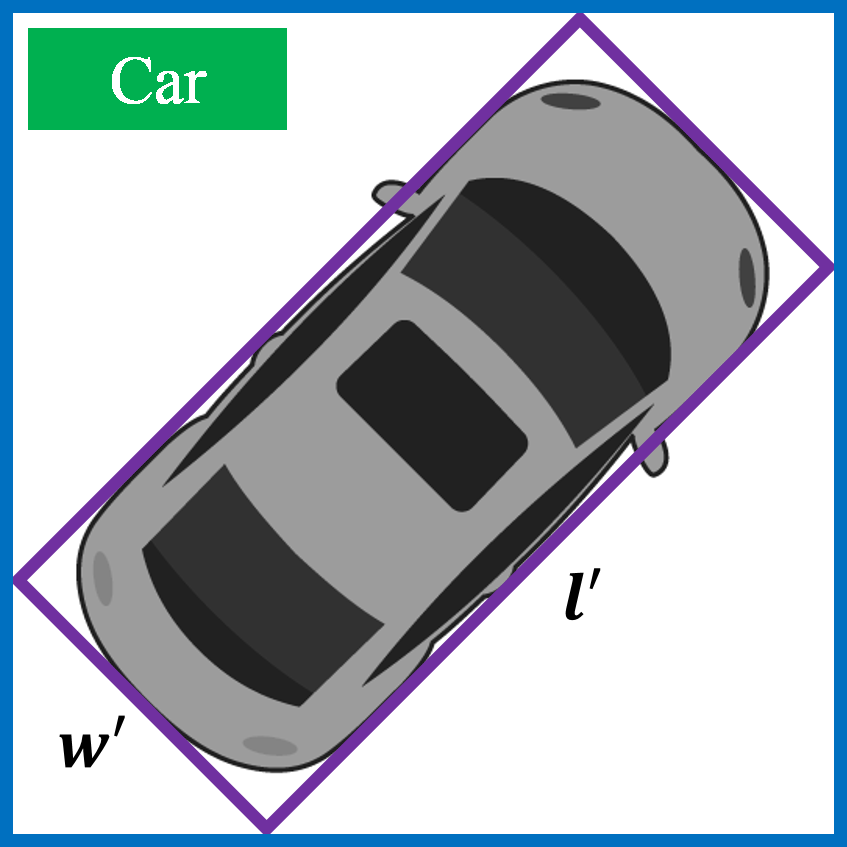}%
\label{fig:2drefinedbox}}
\caption{2D detection refinement process. (a) Network output: class, axis-aligned detection, and yaw estimation. (b) 2D refined box.}
\label{fig:2drefinement}
\end{figure}

Hence, the expected output from the network is a set of bounding boxes, which represent the minimum axis-aligned rectangles enclosing the object-oriented bounding boxes, provided with a category label as well as an estimation of its yaw angle in the ground plane, as shown in Fig. \ref{fig:2ddetection}. 

\subsubsection{Multi-Task Training}
Weights of the CNN are optimized through a multi-task loss that takes into account the different tasks to be accomplished; that is, generation of proposals, classification, and orientation estimation. The latter is treated as a one-of-many classification problem and, therefore, a multinomial logistic loss is used. Different orientation estimations are given for each class, and only the one corresponding to the ground-truth class contributes to the loss. On the other hand, to reduce the impact of class imbalance in the KITTI dataset, we use a weighted multinomial logistic loss where the underrepresented classes have a higher contribution to the final loss.

Weights are initialized with a model pre-trained on ImageNet for recognition on RGB images. As will be shown later, these features prove to be useful to initialize our model, despite the different nature of the input images. Regarding mean subtraction, we assume average values negligible since most cells do not contain information.

\subsubsection{Data Augmentation}
During training, horizontal flipping is used as a data augmentation technique. Additionally, we tackle the fact that KITTI annotations are available only within the camera field of view by generating new samples through the rotation of training BEVs and their annotations 90$^{\circ}$, 180$^{\circ}$ and 270$^{\circ}$ when we train models intended for full 360$^{\circ}$ perception.


\subsection{Post-processing}
In order to obtain accurate 3D object detections, some post-processing needs to be performed over the output of the network.

First of all, the estimated object orientation is used in order to turn the provided axis-aligned detections into object-aligned 2D boxes, as shown in Fig. \ref{fig:2drefinement}. For that purpose, a fixed width $w'$ is used for each object class, based on a statistical analysis of the dimensions of the road users: 1.8m for cars, and 0.6m for pedestrians and cyclists. The two possible object lengths are computed following (\ref{eq:carlength0}) and (\ref{eq:carlength1}), and their corresponding rotated bounding box are obtained. The selected length $l'$ will be the one whose rotated box maximizes the IoU with the axis-aligned detection.

\begin{equation}\label{eq:carlength0}
  l_w = \left|\frac{h_\text{bbox} - \left|\cos (\theta+\frac{\pi}{2}) \cdot w_\text{fixed}\right|}{\cos \theta} \right|
\end{equation}
\begin{equation}\label{eq:carlength1}
  l_h = \left|\frac{w_\text{bbox} - \left|\sin (\theta+\frac{\pi}{2}) \cdot w_\text{fixed}\right|}{\sin \theta} \right|
\end{equation}

Once the oriented 2D boxes are obtained (see Fig. \ref{fig:2drefinedbox}) the height of the objects has to be estimated. Assuming all road users are placed on the ground plane, and considering LiDAR rays may not collide with the lower parts of distant obstacles due to occlusions, a coarse ground plane estimation is performed to get the bottommost coordinate of the obstacle. A grid is created for the XY plane of the laser point cloud, where each of the two-meter-side cells stores the minimum height of the points falling inside. To remove possible noise caused by outliers, a median blur is applied to the resulting grid.

Lastly, the final 3D detection is defined by: the object center $C=(x, y, z)$, where $x$ and $y$ can be obtained from the center of the 2D detection, and $z$ is half the object height $h'$; the object size $S = (l', w', h')$, where $h'$ is the result of subtracting the minimum height, taken from the ground grid cells below the detection, to the maximum height of the object, given by the corresponding channel in the BEV; and the estimated yaw $\theta$ provided by the network.

\section{Experiments}
\label{sec:results} 
The proposed algorithm is evaluated on the KITTI Object Detection Benchmark \cite{Geiger2012KITTI} using the training/validation split from \cite{Chen17}. The performance of our approach is analyzed from different perspectives. Firstly, the effect of different network architecture changes is described. Secondly, the importance of the different channels encoded in the BEV is assessed through a set of ablation studies. Later, a comparison with the most relevant state-of-art methods in the field is performed. Finally, qualitative results showing the multi-device suitability of the method are presented. 

As KITTI annotations are available only in the field of view of the camera, we perform our experiments on a BEV spanning 110$^{\circ}$ in front of the vehicle. Unless otherwise specified, we use a resolution of \SI{0.05}{\metre} per cell and reach \SI{35}{\metre} in front of the vehicle. At test-time, we limit the input to a lateral range of \SI{\pm20}{\metre}. 

\subsection{Architecture Analysis}
As the features needed to perform detection on BEV are necessarily different than those in RGB, we investigate the effect of initializing the weights with a model pre-trained on ImageNet, as it is standard practice in RGB models. Table \ref{tab:randominit} shows the results. Despite the notable differences between both domains, RGB pre-trained weights improve the quality of the feature extractor.

\begin{table}[t]
	\caption{BEV and 3D Detection Performance (\%) Using Different Weight Initialization Strategies, With $N_b=8$.}
	\label{tab:randominit}
	\centering
	\begin{tabular}{ c  c c c  c c c }
		\toprule
        initial weights & \multicolumn{3}{c}{mAP BEV} & \multicolumn{3}{c}{mAP 3D} \\  	
		\cmidrule(lr){2-4} \cmidrule(l){5-7}
		 & Easy & Moder. & Hard & Easy & Moder. & Hard \\ 
		 \midrule   
		ImageNet & 54.46 & 41.61 & 40.57 & 22.92 & 18.02 & 16.92 \\ 
		gaussian & 41.89 & 30.77 & 29.92 & 19.76 & 15.04 & 14.75 \\ 
		\bottomrule
	\end{tabular}
\end{table}

Table \ref{tab:paramselection} shows how variations on different hyper-parameters affect the performance, measured as AP in BEV, on the validation set. We investigate different alternatives:
\begin{enumerate}
\item With/without \textit{pool4}. Removing the fourth pooling layer in the original VGG-16 architecture leads to more resolution on the final feature map, thus improving the performance dramatically on the \textit{Pedestrian} and \textit{Cyclist} categories.
\item With/without ground. Since points belonging to the ground plane apparently do not contribute to the detection task, we tested a setup where those points were removed before generating the BEV image. For this purpose, a height map algorithm was used, where the maximum difference in height between the points inside a cell is computed, and those cells whose values lay below a certain threshold are considered ground plane. Remarkably, this procedure hurt the performance, as the point removal algorithm also deletes points belonging to horizontal surfaces of cars (e.g. the roof). Thus, important information about the objects is lost. 
\item Number of orientation bins, $N_b$. This parameter sets in practice an upper bound for the accuracy of the orientation estimation. However increasing it also presents some problems; e.g. the reduction in the number of training samples per category. We experimented with $N_b=8$ and $N_b=16$, and found little difference between both alternatives. 
\end{enumerate}

From now on, all results correspond to using the no-pool+ground+16bins variant of the network.

\begin{table*}[t]
	\caption{Bird Eye's View Detection Performance (AP BEV) on the Validation Set (\%) for Different Variants.}
	\label{tab:paramselection}
	\centering
	\begin{tabular}{c c c  c c c  c c c  c c c }
		\toprule
		pool4 & ground & $N_b$ & \multicolumn{3}{c}{Car} & \multicolumn{3}{c}{Pedestrian} & \multicolumn{3}{c}{Cyclist}\\  	
		\cmidrule(lr){4-6} \cmidrule(l){7-9} \cmidrule(l){10-12}
		 &  &  & Easy & Moder. & Hard & Easy & Moder. & Hard & Easy & Moder. & Hard \\ 
		 \midrule   
		 No & Yes & 8 					& 72.32 & 54.09 & 54.50 & 43.62 & 39.48 &36.63 & 47.44 & 31.26 &	30.57 \\ 
         No & Yes & 16 		& 73.73 & 54.84 & 56.06 & 44.21 & 39.13 & 35.67 & 50.45 & 33.07 & 31.15 \\ 
         Yes & Yes & 8 		& 70.29 & 49.84 & 54.52 & 25.01 & 23.23 & 21.84 & 41.87 & 27.49 & 25.79 \\
         Yes & No & 8 & 66.63 & 48.52 & 47.98 & 24.59 & 23.07 & 22.25 & 37.60 & 23.55 & 22.62 \\
         No & No & 16 & 70.19 & 52.36 & 52.53 & 41.73 &37.17 & 34.81 & 41.59 & 26.94 & 26.21 \\
        No & No & 8 & 69.80 & 52.56 & 48.44 & 36.19 & 32.97 & 31.39 & 45.23 & 29.32 &	26.89 \\
		\bottomrule
	\end{tabular}
\end{table*}

\subsection{Ablation Studies}
In order to analyze the relevance of the different data stored in the proposed BEV, the three different aforementioned channels have been isolated into individual images, which have been used to train the network separately. As can be seen on Table \ref{tab:ablation}, the least relevant information set for our detection and classification architecture corresponds to intensity values acquired by the LiDAR, as it might be expected due to many factors affecting reflectance measurements \cite{Kashani2015LIDARIntensity}. Besides, both the normalized density and the maximum height channels provide similar numbers, going well beyond those obtained using intensity information. Finally, the results produced when using the three-channel input image exhibit greater AP for all categories, proving their complementarity and, as a consequence, showing the positive effect of aggregating them.

\begin{table*}[t]
	\caption{Bird Eye's View Detection Performance (AP BEV) on the Validation Set (\%) using Different Data as an Input.}
	\label{tab:ablation}
	\centering
	\begin{tabular}{c c c  c c c  c c c  c c c }
		\toprule
		Intensity (R) & Density (G) & Height (B) & \multicolumn{3}{c}{Car} & \multicolumn{3}{c}{Pedestrian} & \multicolumn{3}{c}{Cyclist}\\  	
		\cmidrule(lr){4-6} \cmidrule(l){7-9} \cmidrule(l){10-12}
		 &   &  & Easy & Moder. & Hard & Easy & Moder. & Hard & Easy & Moder. & Hard \\ 
		 \midrule   
          \checkmark &  \checkmark  &  \checkmark  & 72.32 & 54.09 & 54.50 & 43.62 & 39.48 & 36.63 & 47.44 & 31.26 & 30.57 \\
          \midrule  
          \checkmark & &  & 55.04 & 41.16 & 38.56 & 36.25 &	30.43 &	28.37 & 33.09 & 22.83 &	21.79 \\
          & \checkmark &  & 70.94  & 53.00 & 53.30  & 38.21  & 32.72 & 29.58 & 43.77 & 28.62  & 26.99 \\
		  & & \checkmark & 69.80 & 52.90 & 53.69 & 38.37 & 34.04 & 32.37 & 48.06 & 31.21 & 30.40 \\         
		\bottomrule
	\end{tabular}
\end{table*}

\subsection{KITTI Benchmark}

To evaluate the performance of the proposed approach, results over the KITTI test set are provided in Table \ref{tab:comparison_dobem}. For the purpose of a fair analysis, only a comparable method which uses a similar LiDAR projection to provide 3D Detections has been considered.

As can be observed, although both approaches take a BEV image as input, only our method is able to perform detection and classification for different categories. However, it can be noted that this major difference does not prevent BirdNet to rank better in all metrics and every difficulty level, being particularly significant the gap in 3D and BEV, where our method offers more than 2x its performance. In addition, our method is considerably faster.

\begin{figure}[t]
\centering
\subfloat{\includegraphics[width=0.33\linewidth]{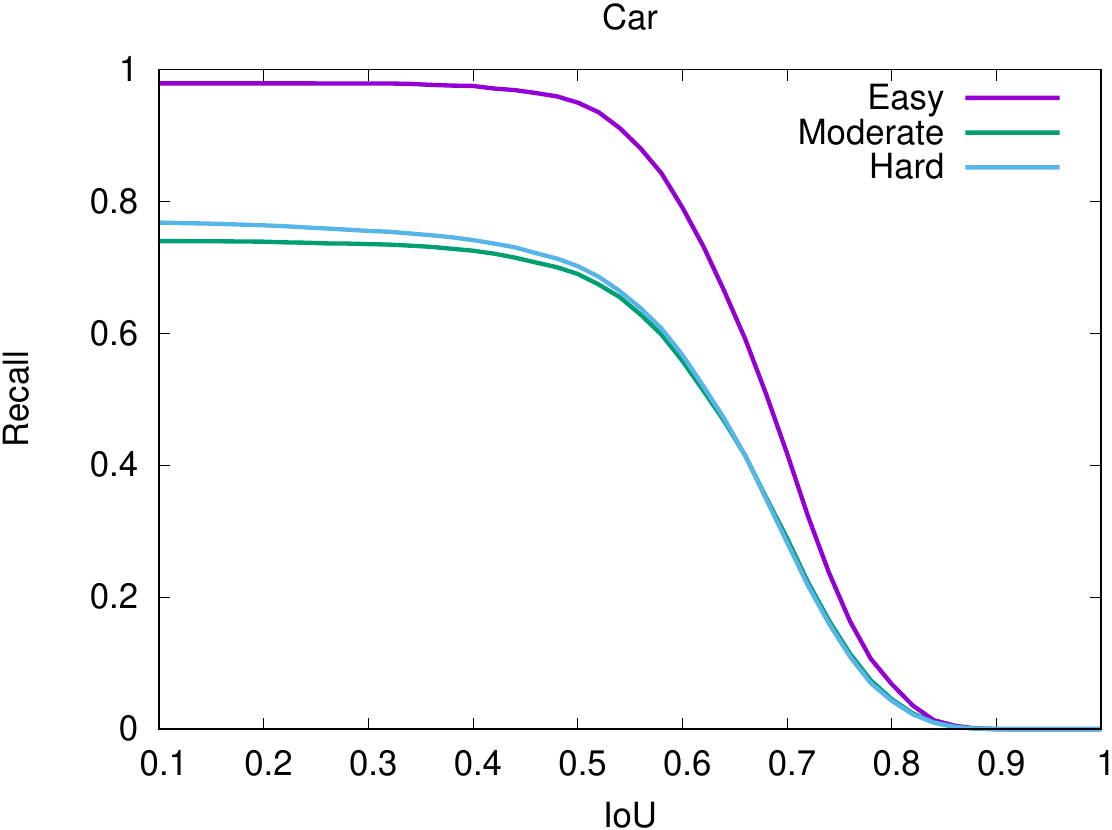}%
\label{fig:car_iou_3d}}
\hfil
\subfloat{\includegraphics[width=0.33\linewidth]{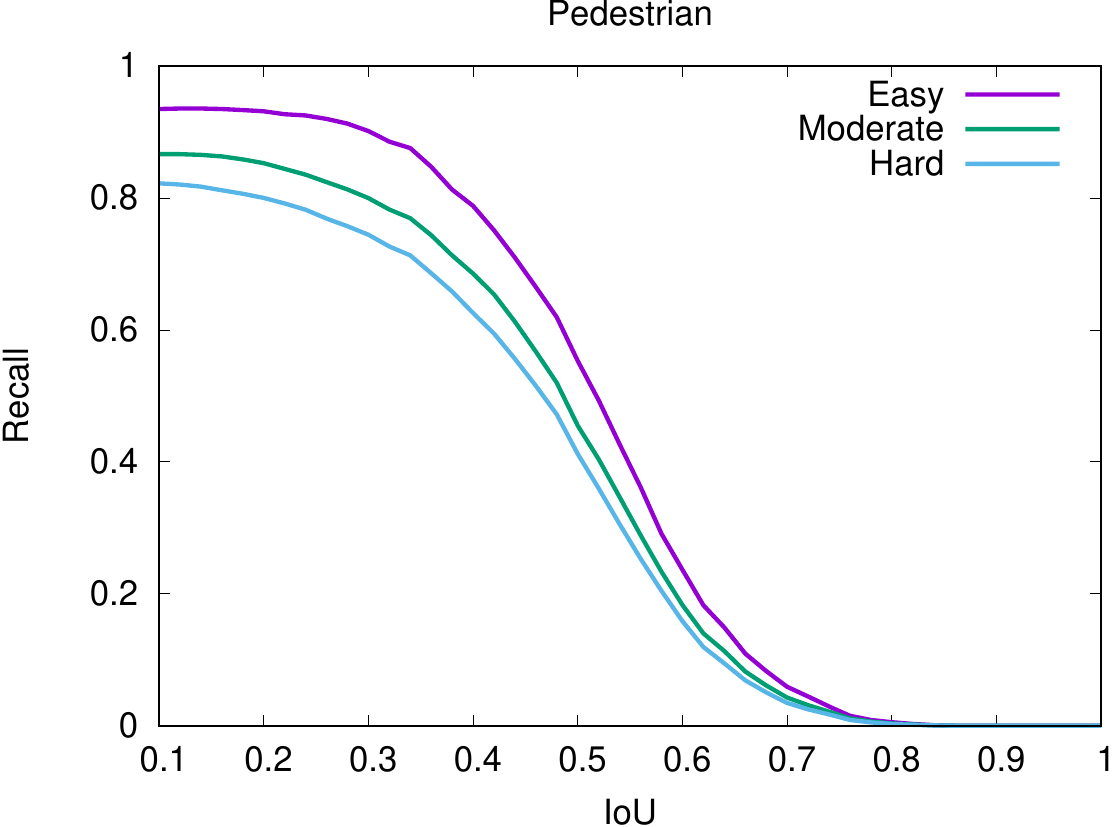}%
\label{fig:ped_iou_3d}}
\hfil
\subfloat{\includegraphics[width=0.33\linewidth]{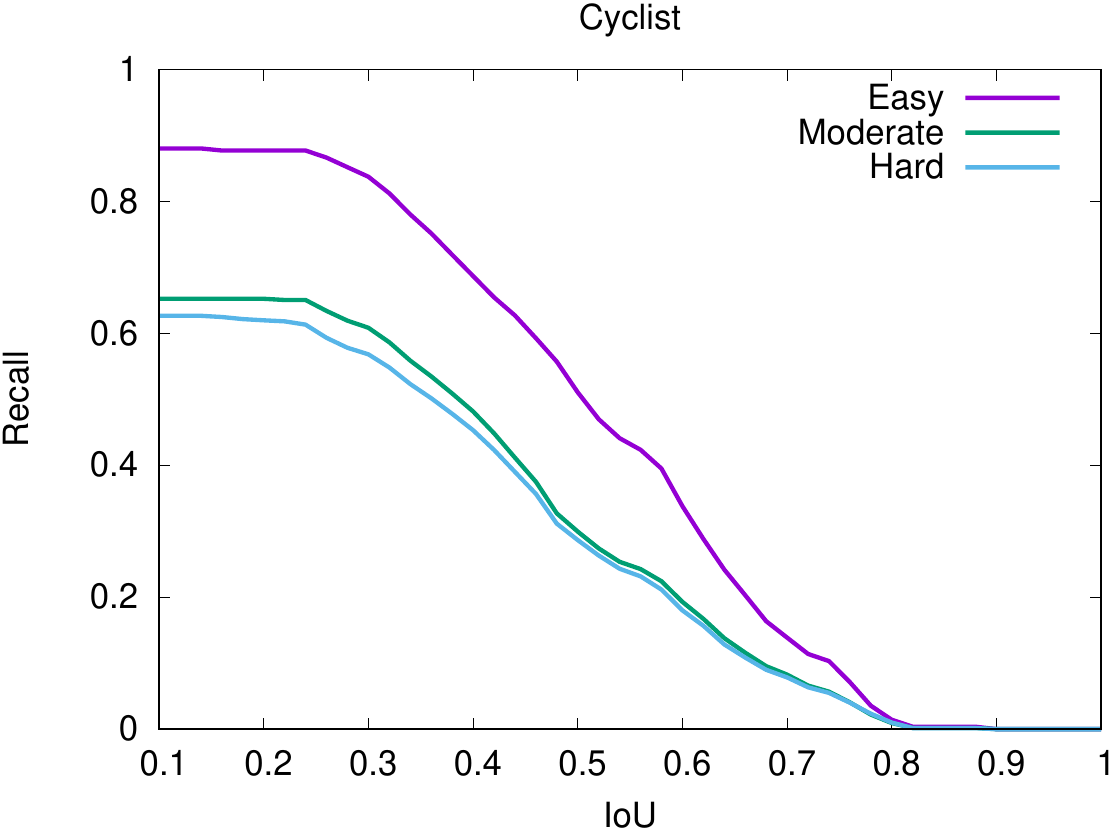}%
\label{fig:cyc_iou_3d}}
\caption{Recall at different IoU thresholds using 300 proposals.}
\label{fig:iourecall}
\end{figure}

Despite these notable results, by studying the recall at different Intersection over Union (IoU) threshold (see Fig. \ref{fig:iourecall} it can be observed that our numbers are heavily affected by the minimum threshold (0.7) required by the KITTI Benchmark. As can be seen, our method is able to locate more than 70\% of vehicles in \textit{Moderate} difficulty taking 0.5 as the minimum IoU. Similarly, our method is capable of detecting pedestrians very efficiently at lower IoU. On the contrary, it has some problems at cyclist detection.

Based on these findings, Table \ref{tab:comparison_0.5} shows a detailed comparison of our method against other state-of-the-art works providing 3D detections. In this case, the \textit{car} category is studied at an $IoU=0.5$, thus only works which have this information published have been considered. Results are given on the validation set, as labels for test set are not publicly available.

As shown in the table, the results provided by our method are remarkable. Compared to VeloFCN, our approach has better performance both in 3D and BEV detection. Regarding the other methods, our numbers are slightly lower in 3D detection and comparable to them in BEV, despite MV(BV+FV) uses different LiDAR projections as inputs, and F-PC\_CNN fuses LiDAR information with RGB. Moreover, it is by far the method with the fastest execution time.

\begin{table*}[t]
	\caption{Comparison of the Bird Eye's View Detection Performance with A Comparable Method on the Test Set (\%).}
	\label{tab:comparison_dobem}
	\centering
	\begin{tabular}{c c c  c c c  c c c  c c c c c c c}
		\toprule
		 Class & Method & \multicolumn{3}{c}{2D detection (AP 2D)} & \multicolumn{3}{c}{2D orientation (AOS)} & \multicolumn{3}{c}{3D detection (AP 3D)} & \multicolumn{3}{c}{BEV detection (AP BEV)} & t (s)\\  	
		\cmidrule(lr){3-5} \cmidrule(l){6-8} \cmidrule(l){9-11} \cmidrule(l){12-14}
		 & & Easy & Moder. & Hard & Easy & Moder. & Hard & Easy & Moder. & Hard & Easy & Moder. & Hard \\ 
		 \midrule   
        \multirow{2}{*}{Car} & DoBem \cite{Yu2017} & 36.35 & 33.61 & 37.78 & 15.35 & 14.02 & 16.33 & 7.42 & 6.95 & 13.45 & 36.49 & 36.95 & 38.10 & 0.6\\ 
        & Ours & 78.18 & 57.47 & 56.66 & 50.85 & 35.81 & 34.90 & 14.75 & 13.44 & 12.04 & 75.52 & 50.81 & 50.00 & 0.11 \\
        \midrule        
        Ped. & Ours &  36.83 & 30.90 & 29.93 & 21.34 & 17.26 & 16.67 & 14.31 & 11.80 & 10.55 & 26.07 & 21.35 & 19.96 & 0.11 \\
        \midrule
        Cyc. & Ours & 64.88 & 49.04 & 46.61 & 41.48 & 30.76 & 28.66 & 18.35 & 12.43 & 11.88 & 38.93 & 27.18 & 25.51 & 0.11 \\
		\bottomrule
	\end{tabular}
\end{table*}

\begin{table*}[t]
	\caption{Comparison of the Bird Eye's View Detection Performance with Other Methods on the Validation Set for Cars with IoU 0.5 (\%).}
	\label{tab:comparison_0.5}
	\centering
    \begin{threeparttable}
	\begin{tabular}{c c  c c c c c c }
		\toprule
		 Method & \multicolumn{3}{c}{3D detection (AP 3D)} & \multicolumn{3}{c}{BEV detection (AP BEV)} & Time(s) \\  	
		\cmidrule(lr){2-4} \cmidrule(l){5-7} 
		 & Easy & Moder. & Hard & Easy & Moder. & Hard   \\ 
		 \midrule   
		MV(BV+FV) \cite{Chen17} & 95.74 & 88.57 & 88.13 & 86.18 & 77.32 & 76.33 & 0.24 \\ 
         VeloFCN \cite{Li16VeloFCN} & 67.92 & 57.57 & 52.56 & 79.68 & 63.82 & 62.80 & 1 \\ 
         F-PC\_CNN (MS)\tnote{*} \cite{Du2018} & 87.16 & 87.38 & 79.40 & 90.36 & 88.46 & 84.75 & 0.5 \\ 
        Ours & 88.92 & 67.56 &	68.59 & 90.43 & 71.45 &	71.34 & 0.11 \\
		\bottomrule
	\end{tabular}
    \begin{tablenotes}
		\item[*] Fuses RGB and LiDAR data.
	\end{tablenotes}
    \end{threeparttable}
\end{table*}

\begin{figure*}[t]
\centering
\subfloat{\includegraphics[height=1.05in]{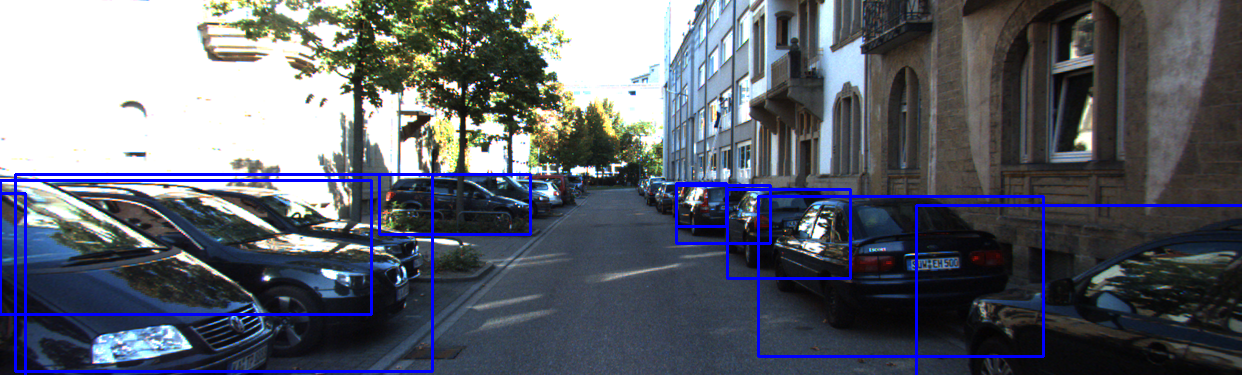}}
\hfil
\subfloat{\includegraphics[height=1.05in]{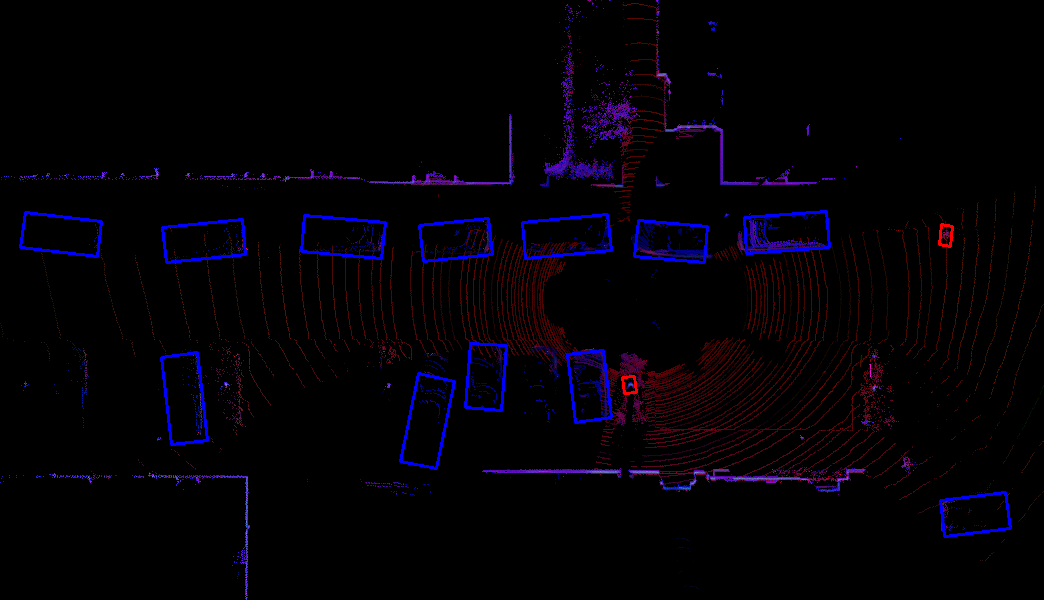}}
\hfil
\subfloat{\includegraphics[height=1.05in]{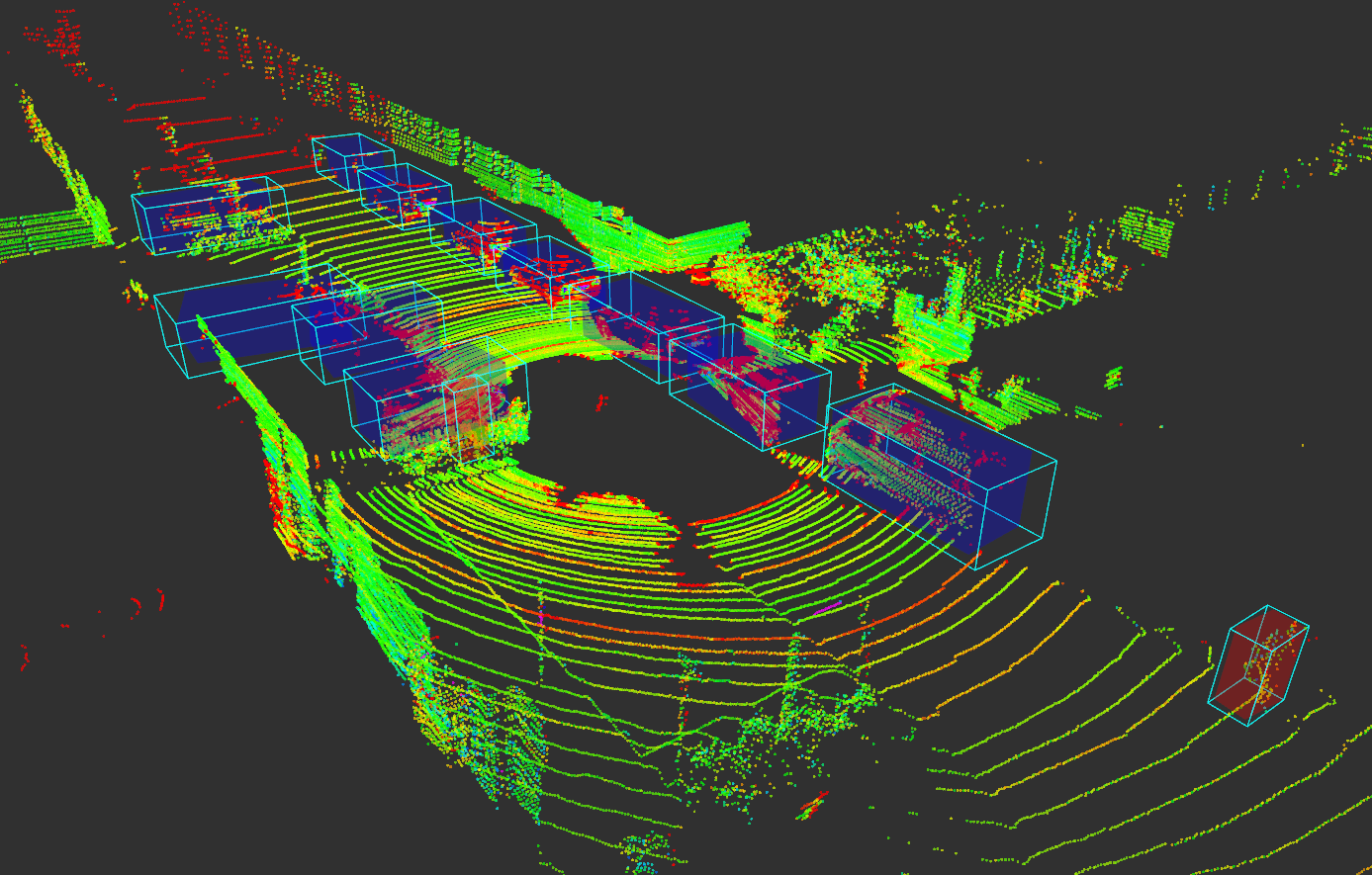}}
\hfil
\subfloat{\includegraphics[height=1.05in]{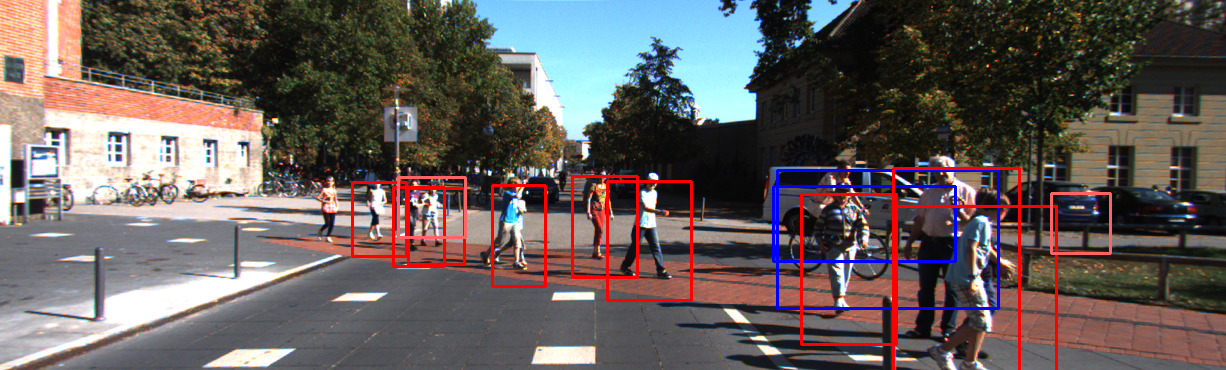}}
\hfil
\subfloat{\includegraphics[height=1.05in]{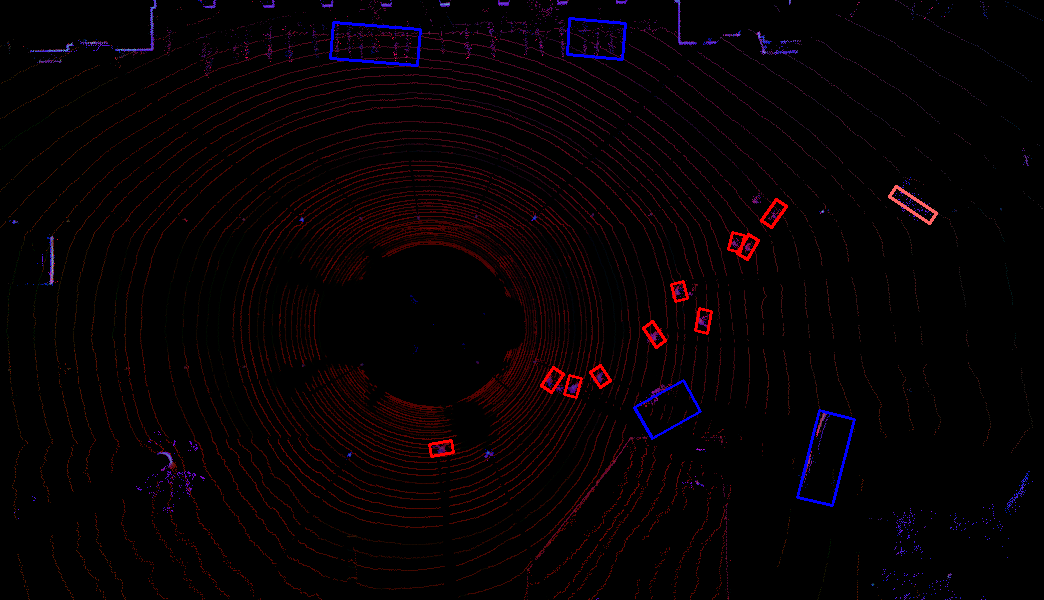}}
\hfil
\subfloat{\includegraphics[height=1.05in]{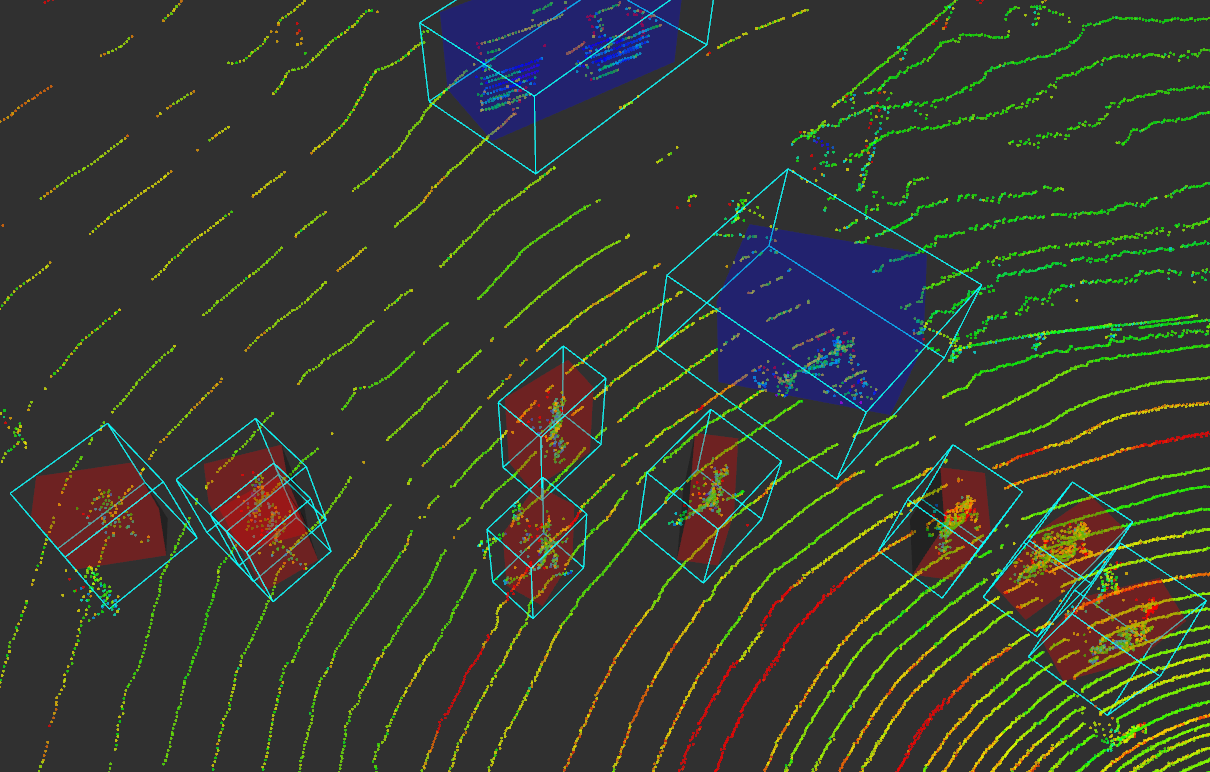}}
\caption{Framework results on KITTI Benchmark test set. From left to right: detections in RGB image, bird eye's view projection, and 3D point cloud.}
\label{fig:test_results}
\end{figure*}

\subsection{Multi-device qualitative results}
To test the robustness of the proposed cell encoding for the BEV projection, experiments using LiDAR devices with different specifications have been conducted. Due to the lack of annotated datasets for these kinds of laser scanners, a unique model has been trained using labels from KITTI Benchmark. Hence, the same weights trained on a 64-layer LiDAR BEV are used for inference in lower resolution inputs. In order to show the effect of the data augmentation techniques described before, qualitative results are given for the full \ang{360} horizontal field of view. The source devices used for these experiments are: Velodyne VLP-16, HDL-32E with 16 and 32 respectively and very diverse vertical angle distributions. Results on HDL-64 data are shown in Fig. \ref{fig:test_results}, as it is the LiDAR sensor used in the KITTI dataset.

\begin{figure}[t]
\centering
\subfloat{\includegraphics[height=1.15in]{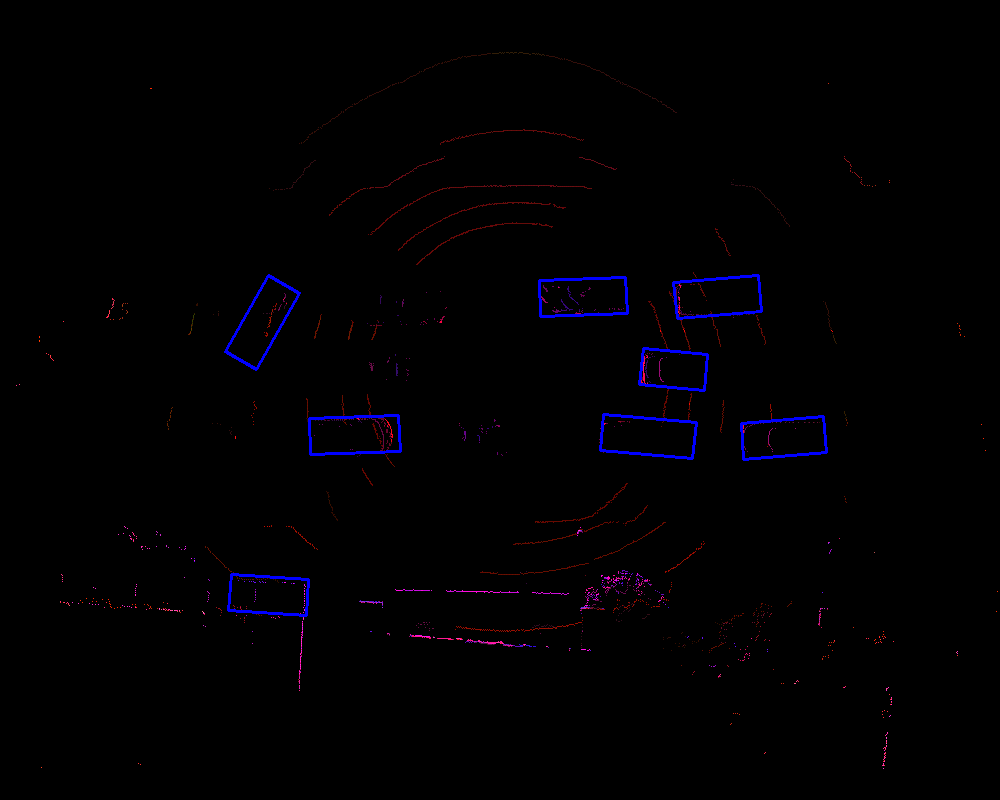}}
\hfil
\subfloat{\includegraphics[height=1.15in]{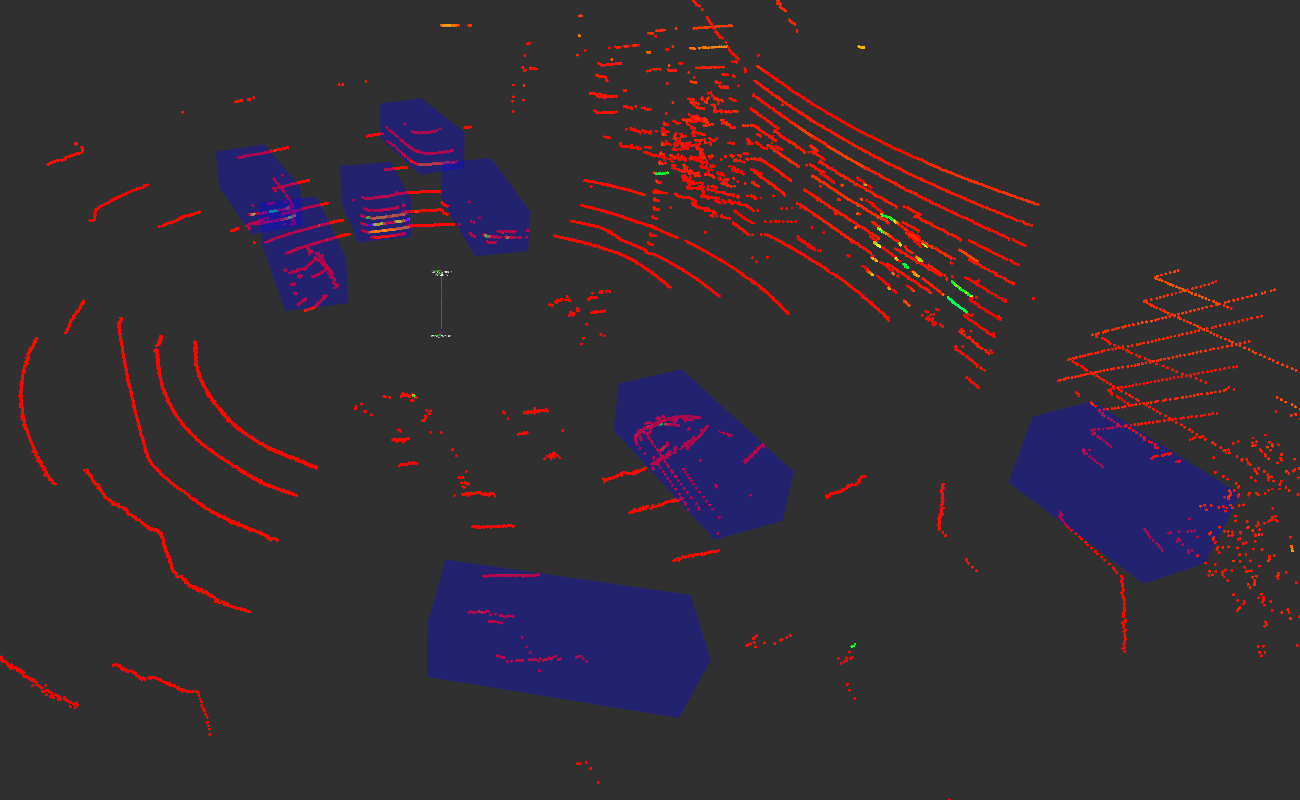}}
\hfil
\subfloat{\includegraphics[height=1.15in]{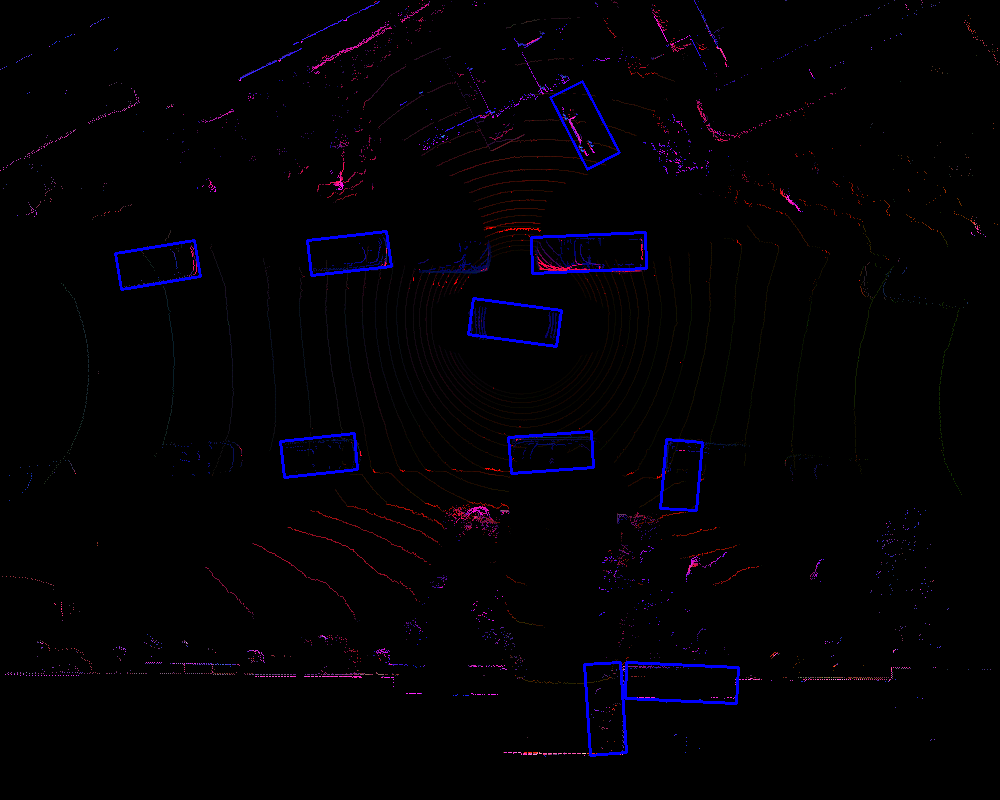}}
\hfil
\subfloat{\includegraphics[height=1.15in]{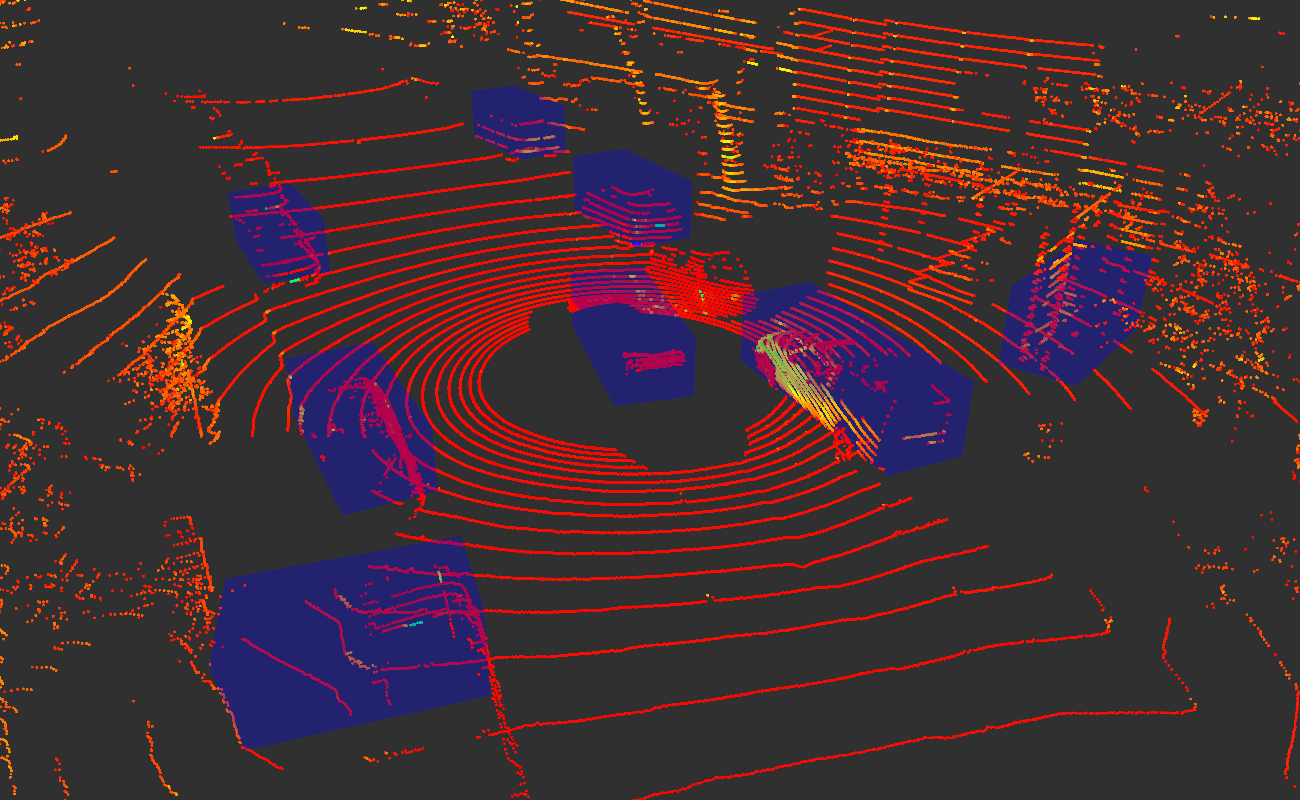}}
\hfil
\caption{Results obtained using different LiDAR devices in \ang{360}. From left to right: 3D estimated boxes in BEV and point cloud. Detections correspond to data from Velodyne VLP-16 (top) and HDL-32E (bottom).}
\label{fig:360_results}
\end{figure}

As can be observed in Fig. \ref{fig:360_results}, a notable performance is showcased on all tested laser inputs, despite their significant differences. Thus, we can state that the information stored on the BEV cells is almost invariant to the sparsity of the original point cloud. However, during experiments, it has been observed that the range at where detections are provided is directly related to the amount of cells containing information, which decreases with the distance. As a result, although the trained network is able to generalize a model for the different categories suitable to be used with different LiDAR devices, the detection range will be proportional to the resolution of the source device.

\section{Conclusions}
\label{sec:conclusions}
In this work, we have presented BirdNet, an efficient approach for 3D object detection in driving environments by adapting a state-of-art CNN framework to process LiDAR data in real-time, making it suitable for on-board operation.

To the best of our knowledge, the approach presented in this paper is the first to introduce pedestrians and cyclist detection using only BEV images as input. According to the results, the proposed detection framework has largely outperformed comparable single-class approaches, both in terms of accuracy and execution time.

Additionally, the assessed multi-device capabilities of the proposed density normalization method, together with the aforementioned data augmentation techniques, will permit training models on already existing high-resolution laser datasets for latter deployment on lower resolution \ang{360} LiDAR setups more appropriate for commercial purposes.

In future work, we plan to improve our method by including 3D anchors proposals into the RPN. Thus, no size assumptions will be required and 3D oriented bounding boxes will be obtained at inference stage. Moreover, we intend to extend the proposed BEV cell encoding by increasing the number of channels to store more information about the original point cloud, such as storing density at different height slices, so more discriminative features can be learned. 

\section*{Acknowledgement} 
Research supported by the Spanish Government through the CICYT projects (TRA2015-63708-R and TRA2016-78886-C3-1-R), and the Comunidad de Madrid through SEGVAUTO-TRIES (S2013/MIT-2713). We gratefully acknowledge the support of NVIDIA Corporation with the donation of the GPUs used for this research. 
\bibliographystyle{IEEEtran}
\bibliography{paper}
\end{document}